\documentclass{article}
\usepackage{spconf,amsmath,graphicx}
\usepackage{subcaption}
\usepackage{graphicscache}

\title{Object-Centric Video Prediction via Decoupling of \\ Object Dynamics and Interactions}
\name{
	Angel Villar-Corrales$^{\dagger}$ \qquad
	Ismail Wahdan$^{\dagger}$ \qquad
	Sven Behnke~\thanks{This work was funded by grant BE 2556/16-2 (Research Unit FOR 2535 Anticipating Human Behavior) of the German Research Foundation (DFG).}}
\address{Autonomous Intelligent Systems, University of Bonn, Germany}
\usepackage{amssymb}
\usepackage{amsmath}
\usepackage{algorithm}
\usepackage[noend]{algpseudocode}
\usepackage{nth}
\usepackage[all]{nowidow}
\usepackage{bbm}
\usepackage{booktabs,etoolbox}
\usepackage{graphicx}
\usepackage{multirow}

\makeatletter
\def\algbackskip{\hskip-\ALG@thistlm}
\makeatother

\usepackage{array}
\newcolumntype{P}[1]{>{\centering\arraybackslash}p{#1}}

\usepackage[export]{adjustbox}

\usepackage[dvipsnames]{xcolor}
\usepackage{enumitem}

\newcommand{\EtAl}{\text{\textit{et al.~}}}

\DeclareMathAlphabet\mathbfcal{OMS}{cmsy}{b}{n}

\usepackage{amsmath}
\DeclareMathOperator*{\softmax}{softmax}

\newcommand{\Images}{\mathbfcal{X}}
\newcommand{\PredImages}{\hat{\mathbfcal{X}}}
\newcommand{\ImageRange}[2]{\mathbfcal{X}_{#1:{#2}}}
\newcommand{\PredImageRange}[2]{\hat{\mathbfcal{X}}_{#1:{#2}}}

\newcommand{\ImageT}[1]{\textbf{X}_{#1}}
\newcommand{\PredImageT}[1]{\hat{\textbf{X}}_{#1}}

\newcommand{\NumContext}{C}
\newcommand{\NumPreds}{T}
\newcommand{\NumObjects}{N}

\newcommand{\SlotDim}{D}
\newcommand{\Slots}{\mathbfcal{S}}

\newcommand{\PredSlots}{\hat{\mathbfcal{S}}}
\newcommand{\SlotsT}[1]{\textbf{S}_{#1}}
\newcommand{\PredSlotsT}[1]{\hat{\textbf{S}}_{#1}}
\newcommand{\SingleSlotsT}[2]{\textbf{s}_{#1}^{#2}}
\newcommand{\SinglePredSlotsT}[2]{\hat{\textbf{s}}_{#1}^{#2}}
\newcommand{\SlotsMany}[2]{\mathbfcal{S}_{#1:#2}}
\newcommand{\PredSlotsMany}[2]{\hat{\mathbfcal{S}}_{#1:#2}}

\newcommand{\SlotsHist}[1]{\textbf{S}^{#1}}

\newcommand{\OutSlotAttention}{\textbf{U}}
\newcommand{\SlotObject}[2]{\textbf{o}_{#1}^{#2}}
\newcommand{\SlotMask}[2]{\textbf{m}_{#1}^{#2}}
\newcommand{\ProcessedMask}[2]{\tilde{\textbf{m}}_{#1}^{#2}}

\newcommand{\Query}{\textbf{Q}}
\newcommand{\Key}{\textbf{K}}
\newcommand{\Value}{\textbf{V}}

\newcommand{\Loss}{\mathcal{L}}

\newcommand{\LossImage}[2]{\mathcal{L}_{I}({#1},{#2})}
\newcommand{\LossObj}[2]{\mathcal{L}_{O}({#1},{#2})}
\newcommand{\LossImageEmpty}{\mathcal{L}_{I}}
\newcommand{\LossObjEmpty}{\mathcal{L}_{O}}

\usepackage{hyperref}
\usepackage{booktabs}                        
\usepackage{tabularx}
\usepackage{multicol}

\def\R{{\mathbb R}}

\newcommand{\Figure}[1]{Fig.~\ref{#1}}

\newcommand{\Table}[1]{Table~\ref{#1}}

\let\svthefootnote\thefootnote
\newcommand\freefootnote[1]{%
	\let\thefootnote\relax%
	\footnotetext{#1}%
	\let\thefootnote\svthefootnote%
}
\begin{document}
	\ninept	
	\maketitle
	\freefootnote{$\dagger$ denotes equal contribution.}
	%
	
	\begin{abstract}
		\vspace*{-0.04cm}
		We present a framework for object-centric video prediction, i.e., parsing a video sequence into objects, and modeling their dynamics and interactions in order to predict the future object states from which video frames are rendered.
		To facilitate the learning of meaningful spatio-temporal object representations and forecasting of their states, we propose two novel object-centric video prediction (OCVP) transformer modules, which decouple the processing of temporal dynamics and object interactions.
		We show how OCVP predictors outperform object-agnostic video prediction models on two different datasets.
		Furthermore, we observe that OCVP modules learn consistent and interpretable object representations.
		Animations and code to reproduce our results can be found in our project website\footnote{\url{https://sites.google.com/view/ocvp-vp}}.
		%
	\end{abstract}


	\section{Introduction}
	\label{sec:intro}
	\vspace{-0.2cm}
	
	Humans perceive the world by parsing scenes into background and multiple foreground objects that can interact with each other~\cite{Johnson_ObjectPerception_2018}.
	Scene modeling approaches that are equipped with inductive biases for such decomposition have the ability to obtain modular and structured representations with desirable properties such as sample efficiency, improved generalization, and more robust representations~\cite{Greff_OnTheBindingProblemInNeuralNetworks_2020, Scholkopf_TowardCausalRepresentationLearning_2021}.
	In recent years, unsupervised approaches to decompose images or video sequences into their object-centric components achieved impressive results on downstream tasks, such as object discovery~\cite{Locatello_ObjectCentricLearningWithSlotAttention_2020, Kipf_ConditionalObjectCentricLearningFromVideo_2022, Burgess_MonetUnsupervisedSceneDecompositionRepresentation_2019} or object tracking~\cite{He_TrackingByAnnimation_2019, Jiang_ScalorGenerativeWorldModelsWithScalableObjectRepresentations_2019}.
	Despite these recent advances in object-centric decomposition, modeling temporal dynamics and object interactions from visual observations alone remains challenging.

	\begin{figure}[t!]
		\centering
		\includegraphics[width=1.0\linewidth]{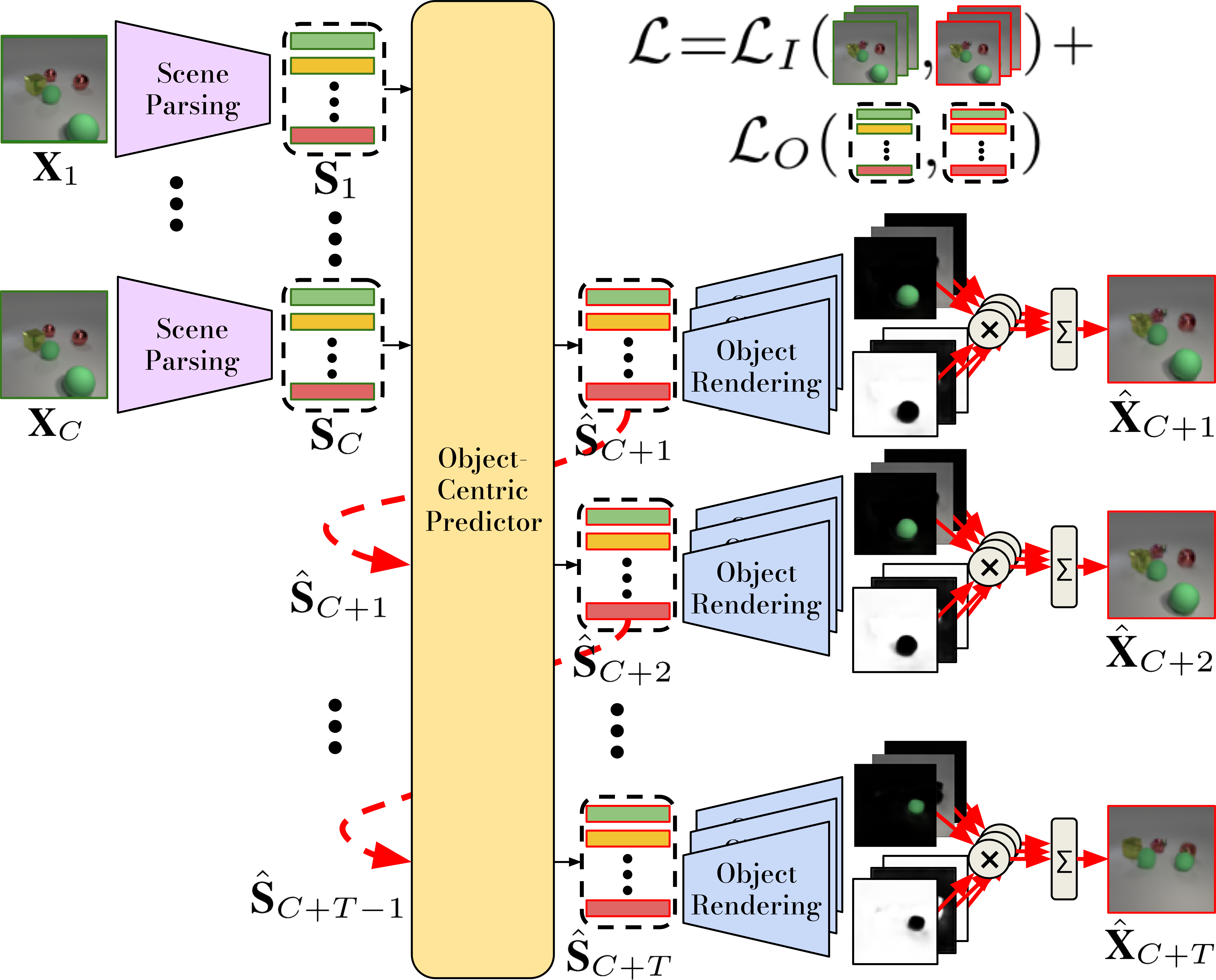}
		\vspace{-0.45cm}
		\caption{
			Overview of our object-centric prediction framework.
			We decompose the seed video frames into object slot representations and learn an autoregressive object-centric predictor to model the object dynamics and interactions so as to predict future object states. 
			Predicted slot representations are independently rendered into object images and masks, which are combined to generate the subsequent video frames.
			Our approach is trained by simultaneously minimizing the slot prediction and video frame prediction mean squared errors.
		}
		\label{fig:teaser}
		\vspace{-0.5cm}
	\end{figure}

	To model spatio-temporal object dynamics, we present a framework for object-centric video prediction.
	Our approach, depicted in \Figure{fig:teaser}, uses a scene parsing module to extract object representations from video frames, learns a sequence prediction module to model temporal dynamics and object interactions using these representations, and renders future video frames from predicted object states.

	A key component in our framework is the sequence predictor, which models multi-object dynamics to predict future object states.
	Different predictor designs embody distinct inductive biases, which affect their suitability for object prediction.
	We investigate multiple predictors for object-centric video prediction and propose two novel object-centric video prediction (OCVP) transformers, which decouple the processing of temporal dynamics and object interactions.

	In summary, our contributions are:
	\textbf{(1)} We present a framework for object-centric video prediction, which decomposes video frames into object representations and models object dynamics and interactions.
	\textbf{(2)} We propose two novel object-centric predictor modules, which decouple the processing of temporal dynamics and object interactions.
	\textbf{(3)} Our prediction framework using our OCVP modules outperforms object-agnostic models for the task of video prediction while learning consistent interpretable object representations.

	\nopagebreak
	\section{Related Work}
	\label{sec:related work}
	
	\hspace{0.0cm} \textbf{Object-Centric Learning:} Our approach is inspired by recent works on unsupervised image and video object-centric decomposition. Image decomposition approaches~\cite{Locatello_ObjectCentricLearningWithSlotAttention_2020, Burgess_MonetUnsupervisedSceneDecompositionRepresentation_2019, Villar_PCDNet_2022, Lin_SPACEUnsupervised_Object-Oriented_SceneRepresentationViaSpatialAttentionAndDecomposition_2020} design models with suitable inductive biases to enforce object-centric decomposition; whereas sequential models~\cite{Kosiorek_SequentialAttendInferRepeat_2018, Kipf_ConditionalObjectCentricLearningFromVideo_2022} leverage the object movement in video sequences to identify objects.
	Our work extends the SAVi~\cite{Kipf_ConditionalObjectCentricLearningFromVideo_2022} video decomposition model to perform object-centric video prediction.
	However, our proposed OCVP modules are general and could be integrated with a variety of decomposition models.

	\vspace{0.15cm}
	
	\hspace{-0.5cm} \textbf{Video Prediction:} Future frame video prediction is the task of forecasting future video frames conditioned on past frames.
	Many different approaches have been proposed for this task, including 3D convolutions~\cite{Gao_SimVPSimplerYetBetterVideoPrediction_2022, Chiu_SegmentingTheFuturue_2020}, RNNs~\cite{Denton_StochasticVideoGenerationWithALearnedPrior_2018, Villar_MSPredVideoPrediction_2022, Wang_PredRNN_2021}, and transformers~\cite{Weissenborn_ScalingAutoregressiveVideoModels_2020, Rakhimov_LatentVideoTransformer_2021}.
	Despite recent advances, most existing methods model temporal changes using image-level features, without explicitly modeling the composition of the video frames or object dynamics.
	A few works follow more structured object-centric approaches for video prediction. Farazi~\EtAl~\cite{Farazi_LocalFrequencyDomainTransformerNwtworksForVideoPrediction_2021} employ local phase differences in order to separate foreground objects from a static background while modeling the foreground motion.
	The approaches presented in~\cite{Wu_GenerativeVideoTransformerCanObjectsBeTheWords_2021, Ye_CompositionalVideoPrediction_2019, Creswell_UnsupervisedObjectBasedTransitionModelsFor3DPartiallyObservableEnvironments_2021, Lin_ImprovingGenerativeImaginationInObjectCentricWorldModels_2020} employ structured or object-centric representations to perform video prediction,	at the cost of requiring explicit human supervision, error-prone Hungarian alignment operations, or being only applicable to very simple 2D datasets.
	The most similar approach to ours, developed concurrently with this work, is SlotFormer~\cite{Wu_SlotFormer_2022}, which also combines SAVi~\cite{Kipf_ConditionalObjectCentricLearningFromVideo_2022} with an autoregressive transformer to perform object-centric video prediction.
	In this work, we further investigate the role of the predictor and propose two novel object-centric transformers that decouple object dynamics and interactions.

	\begin{figure}[t]
		\centering
		\includegraphics[width=.999\linewidth]{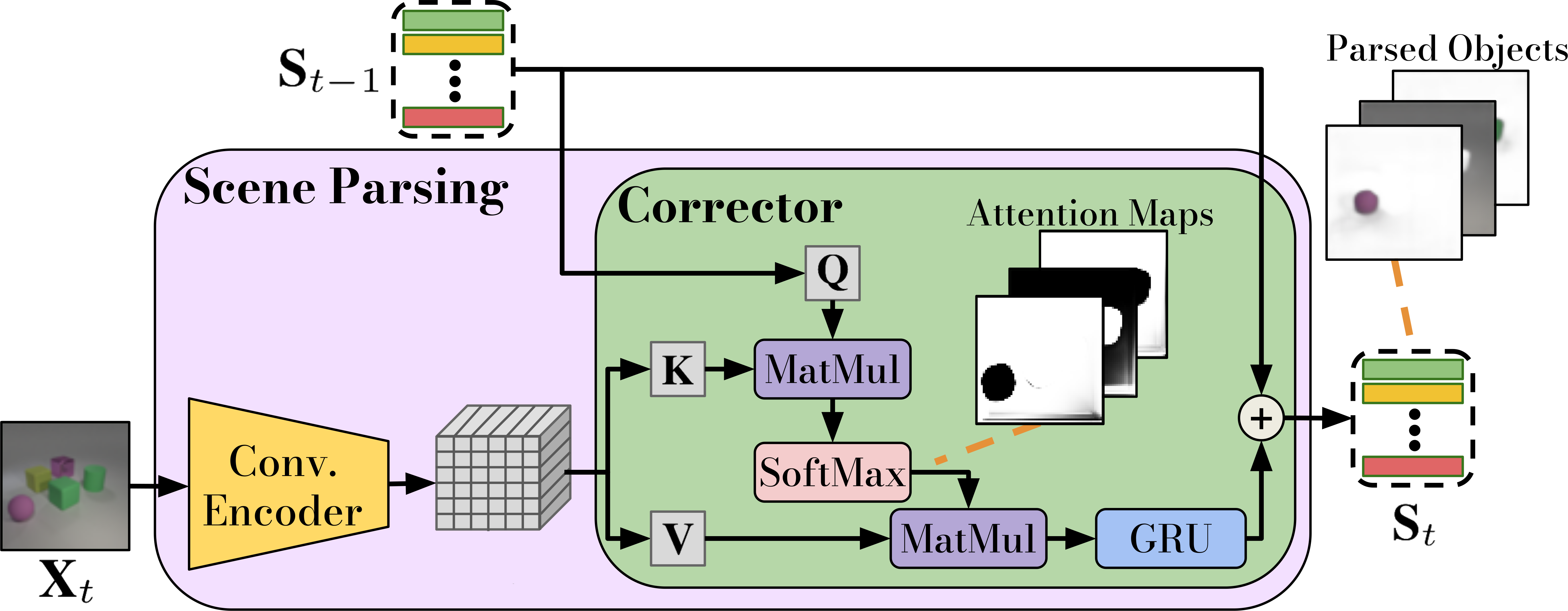}
		\vspace{-0.5cm}
		\caption{
			Scene parsing module.
			We parse seed frames into object slots using a convolutional encoder and a Slot Attention corrector.
		}
		\label{fig: parsing}
		\vspace{-0.4cm}
	\end{figure}

	\section{Method}
	\label{sec:methodology}
	
	Video prediction is defined as the task of, given $\NumContext$ seed video frames $\ImageRange{1}{\NumContext}$, predicting the subsequent $\NumPreds$ frames  $\PredImageRange{\NumContext+1}{\NumContext + \NumPreds}$.
	In this work, we address this task in an object-centric manner, i.e., decomposing the $\NumContext$ seed frames into object representations $\SlotsMany{1}{\NumContext} = (\SlotsT{1}, ..., \SlotsT{\NumContext})$, where $\SlotsT{t}=(\SingleSlotsT{t}{1}, ..., \SingleSlotsT{t}{\NumObjects}) \in \R^{\NumObjects \times \SlotDim}$ is the set of $\SlotDim$-dimensional object embeddings parsed from image $t$, and modeling the object dynamics and interactions to predict future object states and video frames.

	\subsection{Scene Parsing}
	\label{sec:video decomposition}
	
	We employ SAVi~\cite{Kipf_ConditionalObjectCentricLearningFromVideo_2022},
	a recursive encoder-decoder model with a state composed of $\NumObjects$ permutation-invariant object slots,
	to parse the seed video frames $\ImageRange{1}{\NumContext}$ into their object components $\SlotsMany{1}{\NumContext}$, as depicted in \Figure{fig: parsing}.
	The slots $\SlotsT{0}$ are randomly initialized and recursively refined to bind to the objects in the video frames.
	At time step $t$, we encode the input frame into multiple feature maps that represent semantic high-level information of the input.
	These feature maps are fed to a Slot Attention~\cite{Locatello_ObjectCentricLearningWithSlotAttention_2020} corrector, which updates the previous slots based on visual features from the current frame.
	Slot Attention performs cross-attention 
	with the attention coefficients normalized over the slot dimension, thus encouraging the slots to compete to represent parts of the input, and then updates the slot representations using a Gated Recurrent Unit~\cite{Cho_GRU_2014} (GRU).
	Namely, Slot Attention updates the previous slots $\SlotsT{t-1}$ by:
	
	\vspace{-0.2cm}
	\noindent\begin{minipage}{.45\linewidth}
		\begin{align}
			\OutSlotAttention_t = \softmax_{\Query}(\frac{\Query \Key^T}{\sqrt{\SlotDim}}) \Value,
			\nonumber
		\end{align}
	\end{minipage}
	\begin{minipage}{.55\linewidth}
		\begin{equation}
			\SlotsT{t} = \SlotsT{t-1} + \text{GRU}(\OutSlotAttention_t, \SlotsT{t-1}), 
			\label{eq: slot update}
		\end{equation}
	\end{minipage}
	\vspace{-0.1cm}
	
	where $\Key$ and $\Value$ are linear projections of the features maps, and $\Query$ is a linear projection of the previous slots.
	The output of this module is a set of object slots $\SlotsT{t}$, representing the objects of the input frame.

	\begin{figure}[t]
		\centering
		\hspace*{-0.2cm}
		\begin{subfigure}{.167\textwidth}
			\centering
			\includegraphics[width=.858\linewidth]{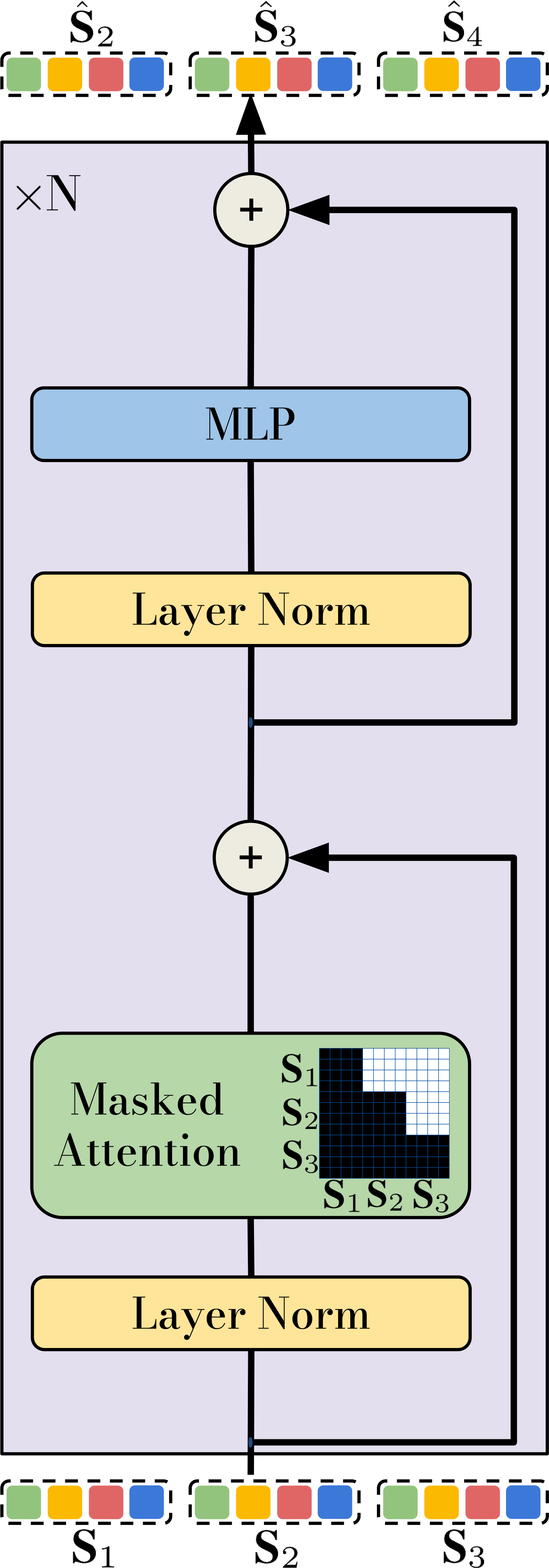}
			\vspace*{-0.1cm}
			\caption{Transformer~\cite{Vaswani_AttentionIsAllYouNeed_2017}}
			\label{fig:VanillaTF}
		\end{subfigure}%
		\hspace*{-0.3cm}
		\begin{subfigure}{.17\textwidth}
			\centering
			\includegraphics[width=.864\linewidth]{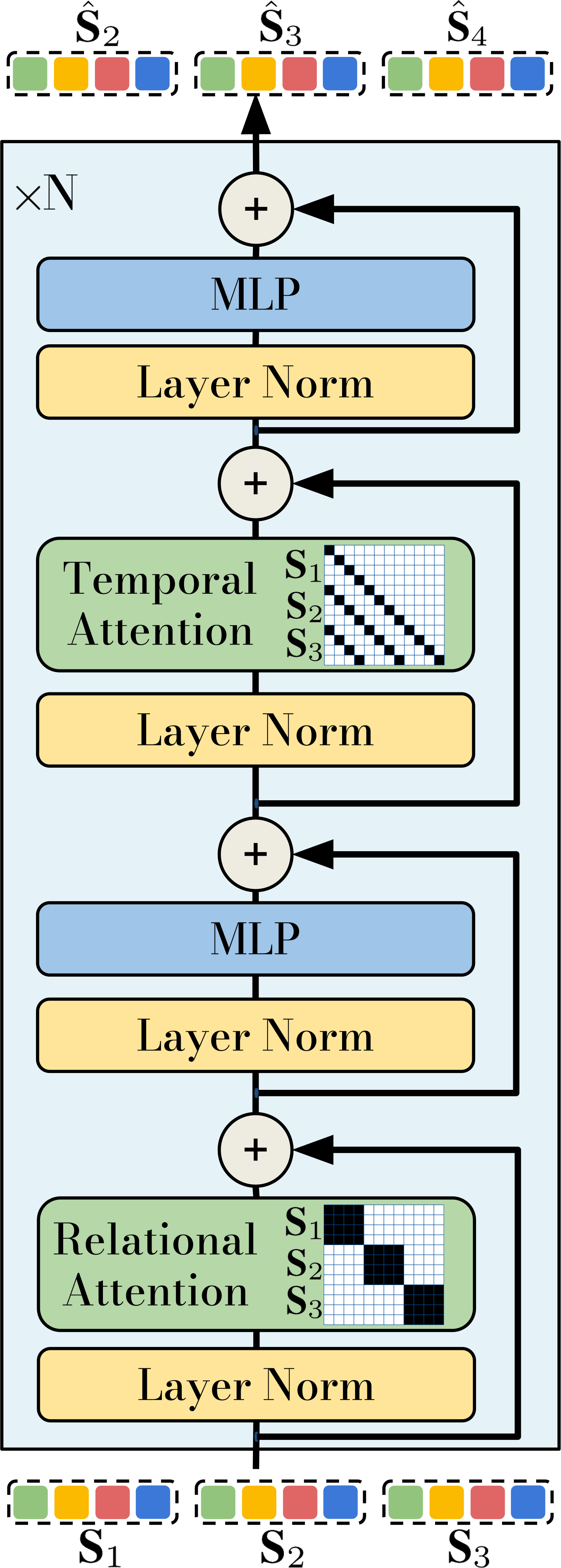}
			\vspace*{-0.1cm}
			\caption{OCVP-Seq}
			\label{fig:OCV-Seq}
		\end{subfigure}%
		\hspace*{-0.2cm}
		\begin{subfigure}{.185\textwidth}
			\centering
			\vspace*{-0.01cm}
			\includegraphics[width=.923\linewidth]{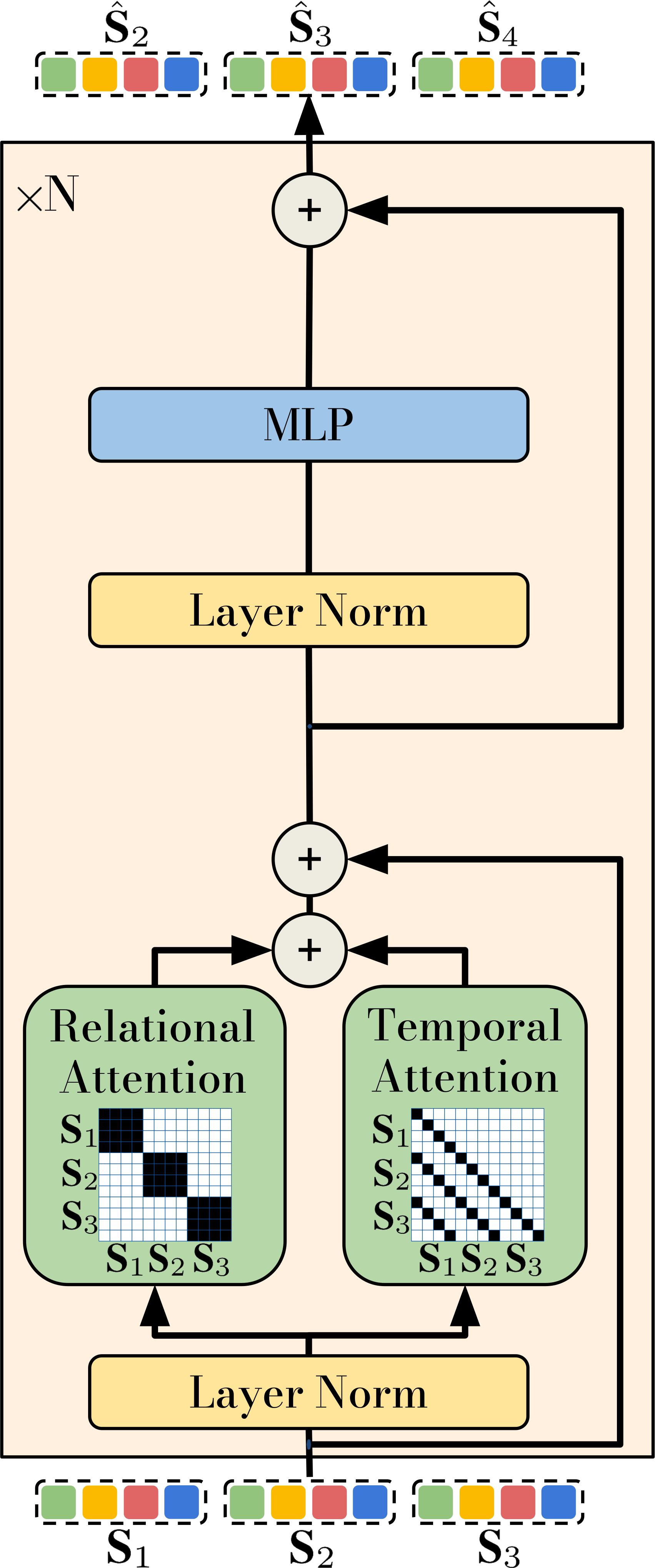}
			\vspace*{-0.1cm}
			\caption{OCVP-Par}
			\label{fig:OCV-Par}
		\end{subfigure}
		\vspace{-0.2cm}
		\caption{
			Transformer predictors employed in our object-centric video prediction framework. The transformer predictor (\ref{fig:VanillaTF}) jointly processes all object slots using masked self-attention, whereas our OCVP modules decouple the processing of temporal dynamics and object interactions using specialized attention blocks, i.e., temporal and relational attention, sequentially (\ref{fig:OCV-Seq}) or in parallel (\ref{fig:OCV-Par}).
		}
		\label{fig:Predictors}
		\vspace{-0.5cm}
	\end{figure}

	\begin{table*}[t!]
		\centering
		\caption{
			Video prediction quantitative evaluation.
			Best two results are highlighted in boldface and underlined.
			We rank the methods according to their performance, averaged across all datasets and metrics.
		}
		\label{table:quantitative comparison}
		\vspace{-0.3cm}
		\footnotesize
		\begin{tabular}{P{1.cm} p{1.8cm} P{0.62cm}P{0.62cm}P{0.62cm} P{0.62cm}P{0.62cm}P{0.75cm} P{0.62cm}P{0.62cm}P{0.62cm} P{0.62cm}P{0.6cm}P{0.8cm} P{0.7cm}}
			\toprule
			&&  \multicolumn{6}{c}{\textbf{Obj3D$_{5 \rightarrow 5}$}} & \multicolumn{6}{c}{\textbf{MOVi-A$_{6 \rightarrow 8}$}} & 
			\multicolumn{1}{c}{}\\
			
			\cmidrule(r){3-8}\cmidrule(r){9-14}
			&& \multicolumn{3}{c}{\textbf{ NumPreds = 15}}
			& \multicolumn{3}{c}{\textbf{ NumPreds = 25}}
			& \multicolumn{3}{c}{\textbf{ NumPreds = 8}}
			& \multicolumn{3}{c}{\textbf{ NumPreds = 18}}
			& \textbf{Mean~}
			\\

			\textbf{} && 
			{ PSNR}$\uparrow$ & {  SSIM}$\uparrow$ & {  LPIPS}$\downarrow$ & {  PSNR}$\uparrow$ & { SSIM}$\uparrow$ & { LPIPS}$\downarrow$ &
			{ PSNR}$\uparrow$ & { SSIM}$\uparrow$ & { LPIPS}$\downarrow$ & { PSNR}$\uparrow$ & { SSIM}$\uparrow$ & { LPIPS}$\downarrow$ & \textbf{Rank}$\downarrow$ \\
			\midrule
			\
			& CopyLast~ & 25.40 & 0.911 & 0.060 & 23.98 & 0.876 & 0.093 & 23.48 & 0.790 & 0.128 & 22.83 & 0.778 & 0.172 & 6.33 \\
			\multirow{2}{1.2cm}{\textbf{Object Agnostic}} & ConvLSTM~\cite{Shi_ConvLSTMNetworkPrecipitationNowcasting_2015} & \textbf{34.32} & 0.929 & 0.046 & 28.95 & 0.849 & 0.112 & 27.12 &  0.877 & 0.248 & 23.85 &\underline{0.831} & 0.337 & 4.83 	\\
			& PhyDNet~\cite{Guen_DisentanglingPhysiscalDynamicsFromUnknownFactorsForVideoPrediction_2020} & 30.08 & 0.921 & 0.034 & 28.21 & \textbf{0.892} & 0.053 &  \textbf{28.36} & \textbf{0.893} & 0.168 & \textbf{26.49 }& \textbf{0.870} & 0.199 & 3.5 	\\
			%
			\midrule
			\multirow{4}{1.2cm}{\textbf{Object Centric}} &LSTM~\cite{Hochreiter_LongShortTermMemory_1997} & 31.13 & 0.900 & 0.039 & 28.83 & 0.849 & 0.071 &  27.82 & 0.827 & 0.089 & 26.14 & 0.784 & 0.133 & 5.17 \\
			& Transformer~\cite{Wu_SlotFormer_2022} & 32.89 & \underline{0.931} & \textbf{0.025} & \underline{30.87} & \textbf{0.892} & \textbf{0.041} & 27.97 & 0.832 & 0.085 & 26.18 & 0.785 & 0.134 & 2.92 	\\
			& OCVP-Seq & \underline{33.10} & \textbf{0.932} & \textbf{0.025} & \textbf{30.93} & 0.891 & \textbf{0.041} &  \underline{27.99} & \underline{0.834} & \textbf{0.082} & 26.24 & 0.789 & \textbf{0.127} & \textbf{1.83} 	\\
			&OCVP-Par & 32.99 & \underline{0.931} & \textbf{0.025} & 30.85 & 0.890 & 0.043 &  \underline{27.99} & 0.832 & \underline{0.083} & \textbf{26.31} & 0.788 & \textbf{0.127} & \underline{2.67} 	\\
			\bottomrule
		\end{tabular}
		\vspace{-0.4cm}
	\end{table*}

	\subsection{Object-Centric Video Predictor Modules}
	\label{sec:predictor modules}
	
	Given the object slots $(\SlotsT{1}, ..., \SlotsT{\NumContext})$,
	the object-centric predictor module models the temporal dynamics and object interactions to forecast the subsequent object states $\PredSlotsT{\NumContext + 1}$ while maintaining temporally consistent and interpretable object representations.

	To this end, we propose two novel object-centric video prediction (OCVP) transformer modules, depicted in \Figure{fig:Predictors}, which are explicitly designed to decouple the modeling of the temporal and object dimensions by employing two specialized multi-head self-attention mechanisms, i.e., \emph{temporal attention} and \emph{relational attention}.

	\vspace{0.03cm}
	\hspace{-0.cm} \textbf{Temporal Attention:} In the temporal attention block, each object slot is updated by aggregating information from that same object up to the current time step, without modeling interactions between distinct objects.
	Given object slots $(\SlotsT{1}, ..., \SlotsT{\NumContext})$, these are rearranged into temporal histories $(\SlotsHist{1}, ..., \SlotsHist{\NumObjects})$, where $\SlotsHist{n} = (\SingleSlotsT{1}{n}, ..., \SingleSlotsT{\NumContext}{n}) \in \R^{\NumContext \times \SlotDim}$ denotes the temporal history of the $n$-th object slot up to the current time step $\NumContext$.
	Following the transformer design~\cite{Vaswani_AttentionIsAllYouNeed_2017},
	slots in the history are first linearly mapped into key $\Key$, query $\Query$ and value $\Value$ embeddings via learnable projections and then jointly processed via dot-product attention:	
	
	\vspace{-0.5cm}
	\begin{align}
		\text{Attention}(\Query, \Key, \Value) = \softmax_{\Key}(\frac{\Query \Key^T}{\sqrt{\SlotDim}}) \Value  ,
	\end{align}
	\vspace{-0.4cm}
	
	whose outcome yields slot representations that have been updated according to their temporal history.

	\vspace{0.03cm}
	\hspace{-0.cm} \textbf{Relational Attention:} In contrast, relational attention models the multi-object interactions and relationships by jointly processing all slots from the same time step.
	The computation of this module is analogous to the one of temporal attention, with the notable difference that, instead of processing temporal histories from each independent object, it jointly processes all object slots from the same time step.
	The outcome of this block yields object slots that have been updated according to their relations with other objects.

	\vspace{0.02cm}
	\hspace{-0.cm} \textbf{OCVP Modules:} Employing our structured attention blocks, we propose two distinct OCVP modules.
	\emph{OCVP-Seq} (\Figure{fig:OCV-Seq}) cascades both attention blocks, thus iteratively refining the slots with relational and temporal information, respectively;	
	whereas \emph{OCVP-Par} (\Figure{fig:OCV-Par}) applies both attention blocks in parallel and sums their outcomes to aggregate the temporal and object information.

	\subsection{Video Rendering}
	\label{sec:video rendering}
		
	We use the SAVi decoder to independently render an object image and a mask from each individual predicted slot.
	The object masks are normalized across slots using a softmax, and then combined with the rendered objects via a weighted sum to generate video frames:

	\vspace{-0.2cm}
	\noindent\begin{minipage}{.31\linewidth}
		\begin{align}
			\SlotObject{t}{n}, \SlotMask{t}{n} = f_{\text{dec}}(\SinglePredSlotsT{t}{n}), 
			\nonumber
		\end{align}
	\end{minipage}
	\begin{minipage}{.31\linewidth}
		\begin{equation}
			\ProcessedMask{t}{n} = \softmax_{\NumObjects}(\SlotMask{t}{n}),  \nonumber
		\end{equation}
	\end{minipage}
	\begin{minipage}{.38\linewidth}
		\begin{equation}
			\PredImageT{t} = \sum_{n=1}^{N} \SlotObject{t}{n} \odot \ProcessedMask{t}{n}.
		\end{equation}
	\end{minipage}
	\vspace{-0.4cm}

	\subsection{Model Training and Inference}
	\label{sec:model training and inference}
	
	Our object-centric video prediction approach is trained in two separate stages.
	First, SAVi is trained to parse video frames into their object components.
	Then, given the pretrained SAVi model, we train the predictor to forecast future object states.

	More precisely, we parse the seed frames $\ImageRange{1}{\NumContext}$ into their object components $\SlotsMany{1}{\NumContext}$ using the pretrained SAVi encoder.
	Then, the extracted object slots are fed to the predictor module in order to forecast the subsequent object states $\PredSlotsT{\NumContext + 1}$, which are rendered into object images and masks using the pretrained SAVi decoder and then combined to generate the subsequent video frame $\PredImageT{\NumContext + 1}$.
	We apply this process in an autoregressive way to predict, conditioned on all previous parsed and predicted object slots, the subsequent $\NumPreds$ object slots $\PredSlotsMany{\NumContext + 1}{\NumContext + \NumPreds}$ and video frames $\PredImageRange{\NumContext + 1}{\NumContext + \NumPreds}$.

	We train the predictor by minimizing the combined loss:

	\vspace{-0.4cm}
	\begin{align}
		& \Loss = \LossImage{\Images}{\PredImages} + \LossObj{\Slots}{\PredSlots},  \label{eq: loss} \\
		& \LossImageEmpty = \frac{1}{\NumPreds} \sum_{t=1}^{\NumPreds} || \PredImageT{\NumContext + t} - \ImageT{\NumContext+t}||_2^2 \label{eq: image loss},  \\
		& \LossObjEmpty  =  \frac{1}{\NumPreds \cdot \NumObjects} \sum_{t=1}^{\NumPreds}  \sum_{n=1}^{\NumObjects} || \SinglePredSlotsT{\NumContext + t}{n} - \SingleSlotsT{\NumContext + t}{n} ||_2^2 \label{eq: slot loss} , 
	\end{align}
	\vspace{-0.3cm}

	where $\LossImageEmpty$ measures the future frame prediction error, and $\LossObjEmpty$ measures the error when predicting future object slots.
	%

	\section{Experiments}
	\label{sec:experiments}

\begin{table}[t!]
	\centering
	\caption{
		Object-centric prediction evaluation on MOVi-A.
	}
	\label{table:object evaluation}
	\vspace{-0.2cm}
	\footnotesize
	\begin{tabular}{p{2.cm} P{0.62cm}P{0.75cm} P{0.62cm}P{0.75cm} P{0.7cm}}
		\toprule
		
		& \multicolumn{2}{c}{\hspace{-0.2cm}  \textbf{ NumPreds = 8}}
		& \multicolumn{2}{c}{\hspace{-0.2cm}  \textbf{ NumPreds = 18}}
		& \multicolumn{1}{c}{\hspace{-0.2cm}  \textbf{ Mean }}
		\\
		\textbf{} & {ARI}$\uparrow$ & {mIoU}$\uparrow$  &  {ARI}$\uparrow$ & {mIoU}$\uparrow$ & \textbf{Rank}$\downarrow$ \\
		\midrule
		LSTM~\cite{Hochreiter_LongShortTermMemory_1997} & 0.619 & 0.578 & 0.431 & 0.500 & 4 \\
		Transformer~\cite{Wu_SlotFormer_2022} & \textbf{0.640} & 0.585 & \textbf{0.452} & 0.508 & 2 \\
		OCVP-Seq & 0.632 & \textbf{0.588} & \underline{0.443} & \textbf{0.511} & \textbf{1.75} \\
		OCVP-Par & \underline{0.636} & \underline{0.586} & 0.435 & \textbf{0.511} & 2 \\
		\bottomrule
	\end{tabular}
	\vspace{-0.4cm}
\end{table}

\begin{figure*}[t!]
	\begin{subfigure}{.5\linewidth}
		\centering
		\vspace*{-0.0cm}
		\includegraphics[width=0.999\linewidth]{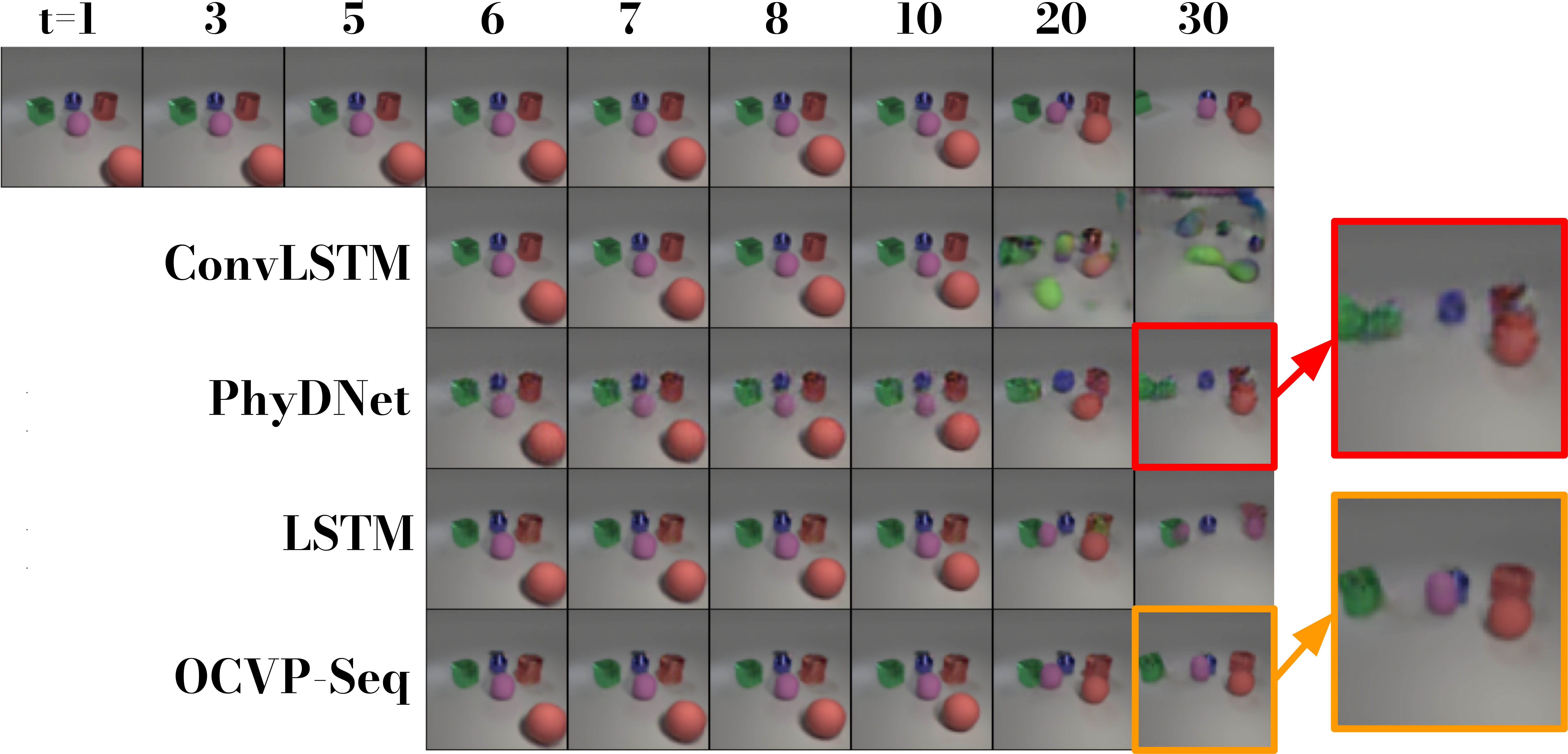}
		\begin{flushleft}
		\vspace{-0.7cm}
		\begin{minipage}[h]{0.1\linewidth}
			\caption{}
			\label{fig:qual obj}
		\end{minipage}
		\end{flushleft}
	\end{subfigure}
	\begin{subfigure}{.5\textwidth}
		\centering
		\vspace*{-0.0cm}
		\includegraphics[width=0.999\linewidth]{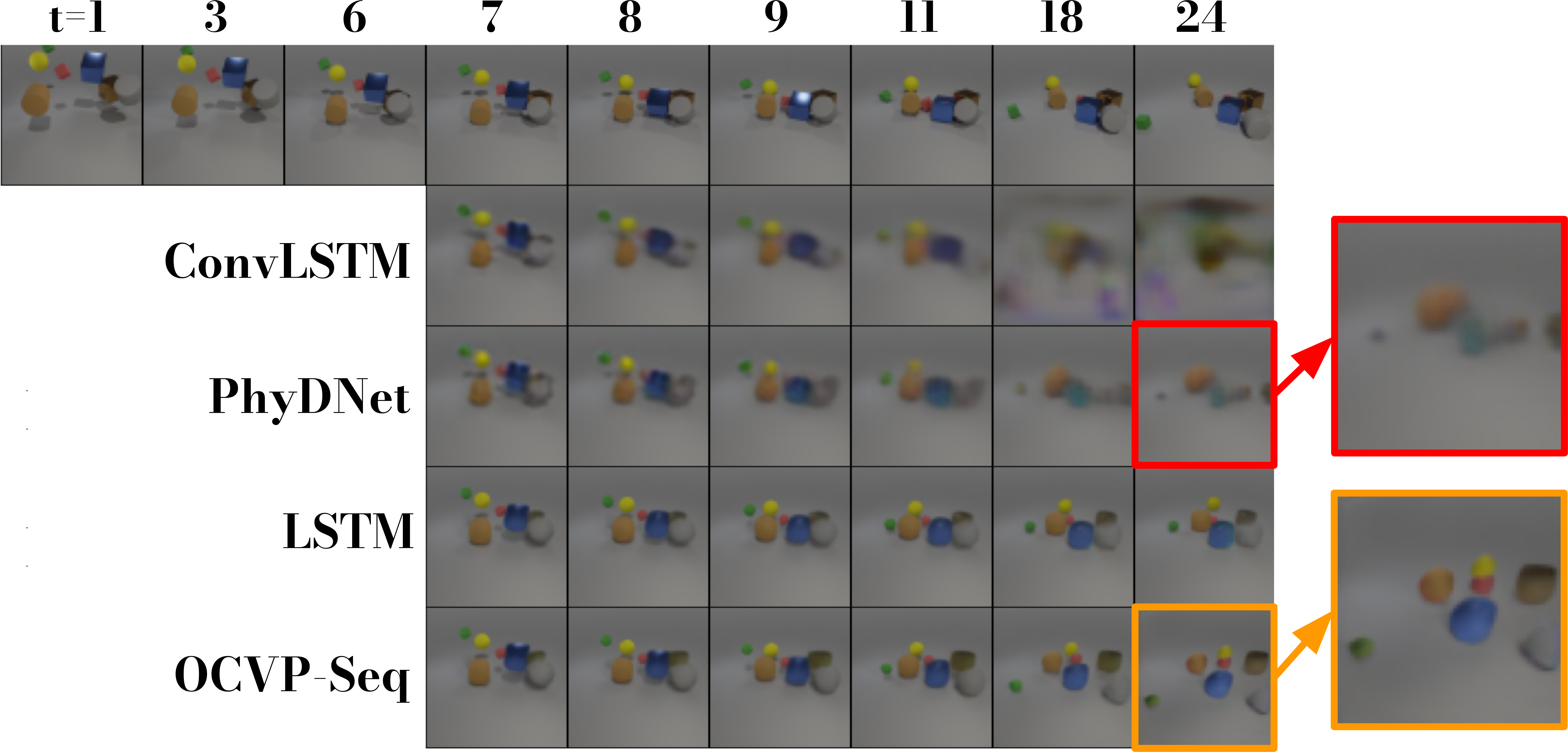}
		\begin{flushleft}
		\vspace{-0.7cm}
		\begin{minipage}[h]{0.1\linewidth}
			\caption{}
			\label{fig:qual movi}
		\end{minipage}
		\end{flushleft}
	\end{subfigure}
	\vspace{-0.3cm}
	\caption{
		Qualitative comparison on Obj3D (\ref{fig:qual obj}) and MOVi-A (\ref{fig:qual movi}).
		We display three seed and six target frames (top row), as well as predictions from four different models.
		OCVP-Seq achieves the most accurate predictions among the compared methods.
		%
	}
	\label{fig:qualitative}
	\vspace{-0.5cm}
\end{figure*}

\begin{figure}[t!]
	\begin{subfigure}{.999\linewidth}
		\centering
		\vspace*{-0.0cm}
		\includegraphics[width=0.999\linewidth]{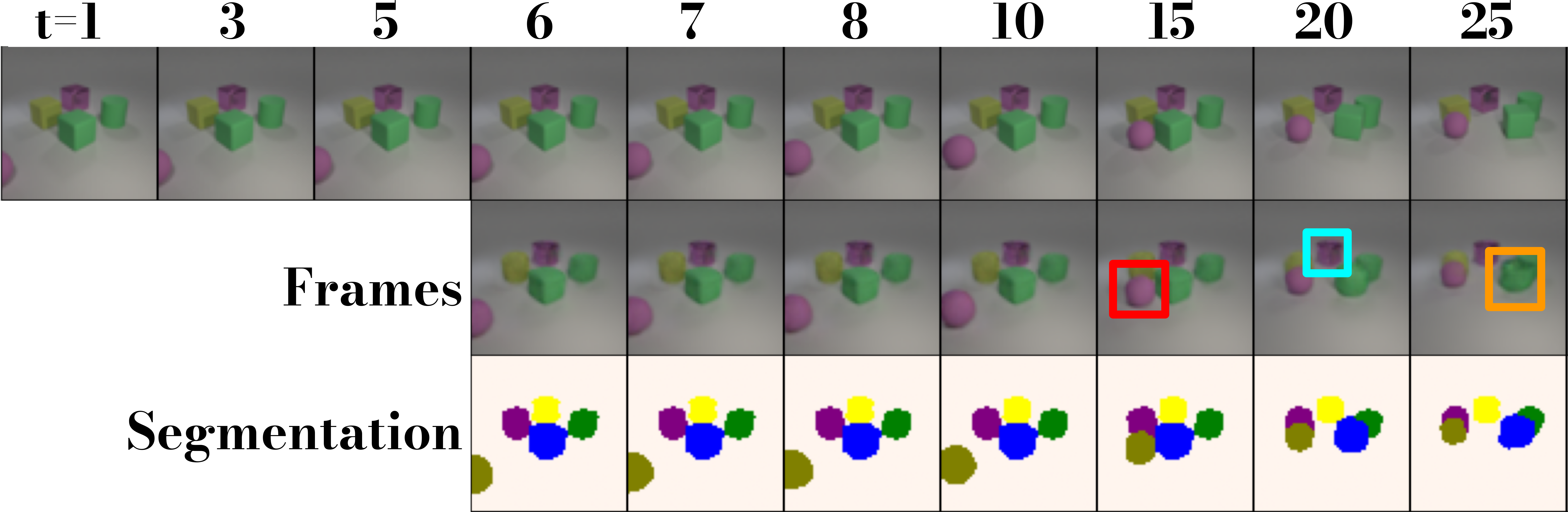}
		\begin{flushleft}
			\vspace{-0.6cm}
			\begin{minipage}[h]{0.06\linewidth}
				\caption{}
				\label{fig: segmentation}
			\end{minipage}
		\end{flushleft}
	\end{subfigure}
	\\
	\begin{subfigure}{.999\linewidth}
		\vspace*{0.1cm}
		\includegraphics[width=0.999\linewidth]{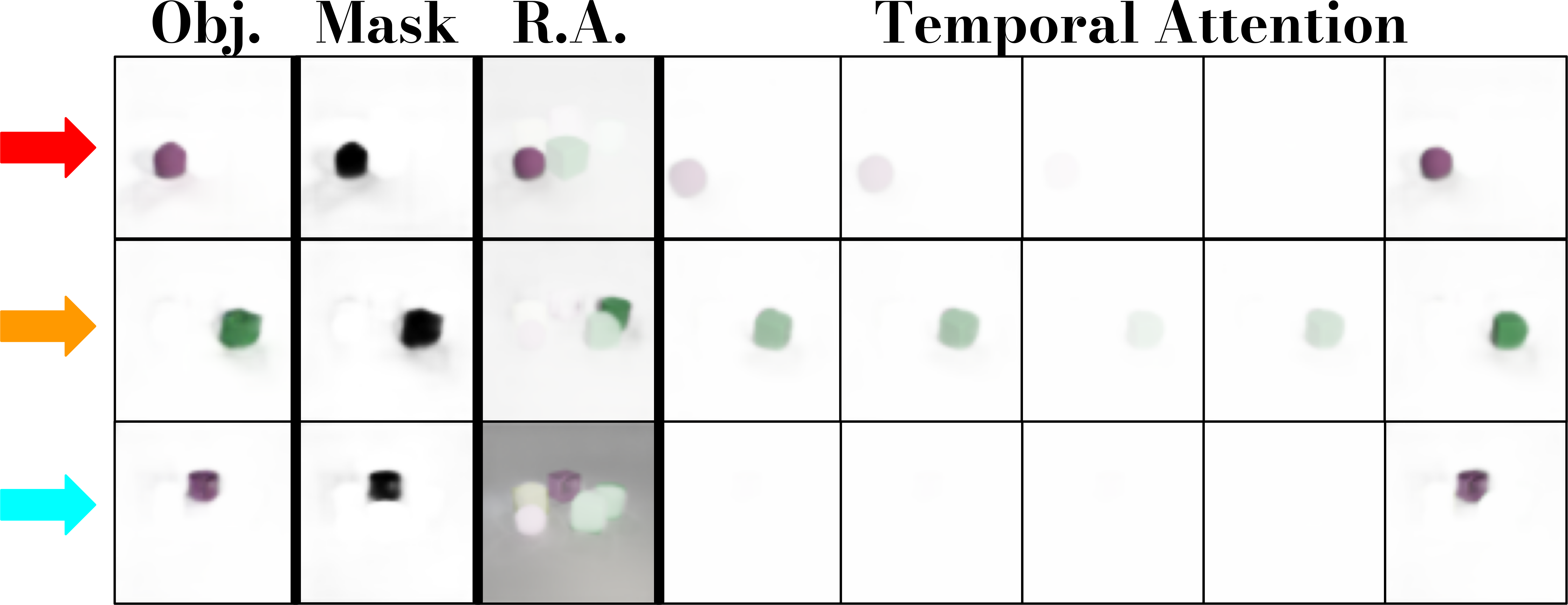}
		\begin{flushleft}
			\vspace{-0.6cm}
			\begin{minipage}[h]{0.06\linewidth}
				\caption{}
				\label{fig: representations}
			\end{minipage}
		\end{flushleft}
	\end{subfigure}
	\vspace*{-0.4cm}
	\caption{
		a) Qualitative segmentation forecasting results on Obj3D using OCVP-Seq.
		b) Interpretable representations for three object predictions. We visualize the predicted object images and masks, and their relational (R.A.) and temporal attention masks.
	}
	\label{fig:segmentation}
	\vspace{-0.4cm}
\end{figure}

	\subsection{Datasets and Experimental Details}
	\label{sec:datasets}
	
	We evaluate our object-centric video prediction framework on two distinct object-centric datasets, namely Obj3D and MOVi-A.
	
	\textbf{Obj3D~\cite{Lin_ImprovingGenerativeImaginationInObjectCentricWorldModels_2020}} contains 2920 train and 200 test synthetic sequences in which a moving ball collides with static 3D geometric objects (e.g. spheres or cubes), setting them in motion.
	We train our predictors using $\NumContext=5$ seed frames to predict the subsequent $\NumPreds=5$ frames.
	
	\textbf{MOVi-A~\cite{Kipf_ConditionalObjectCentricLearningFromVideo_2022}} contains 9703 train and 250 test sequences with up to ten objects similar to those in Obj3D, but with more complex dynamics, occlusions and collisions.
	We train our predictors using $\NumContext=6$ seed frames to predict the subsequent $\NumPreds=8$ video frames.
	
	Following previous works~\cite{Wu_SlotFormer_2022, Locatello_ObjectCentricLearningWithSlotAttention_2020}, we evaluate the visual quality of predicted video frames using  video prediction metrics (PSNR, SSIM and LPIPS), and evaluate the ability to model object dynamics by measuring the ARI and mIoU between ground-truth instance segmentation and the forecasted object masks.

	\subsection{Evaluation}
	\label{sec:evaluation}

	We evaluate our object-centric prediction framework using different predictor modules, namely LSTM~\cite{Hochreiter_LongShortTermMemory_1997}, Transformer~\cite{Wu_SlotFormer_2022}, and our two proposed OCVP modules, for different prediction horizons (\emph{NumPreds}).
	We compare our object-centric approach with two existing object-agnostic prediction models: ConvLSTM~\cite{Shi_ConvLSTMNetworkPrecipitationNowcasting_2015} and PhyDNet~\cite{Guen_DisentanglingPhysiscalDynamicsFromUnknownFactorsForVideoPrediction_2020}.
	Additionally, we include a CopyLast baseline that naively copies the last seed frame.

	\hspace{-0.cm} \textbf{Video Prediction:}
	In \Table{table:quantitative comparison}, we quantitatively report the visual quality of the predicted frames.
	We aggregate scores by measuring how each model ranks compared to the others, and average ranks across metrics and prediction horizons.
	Our OCVP modules achieve the overall best performance, with OCVP-Seq widely outranking all other models, and outperform the object-agnostic baselines and other predictors.	
	This is most notable on the perceptual LPIPS metric, where our OCVP modules achieve perceptually realistic predictions, whereas object-agnostic baselines are even outperformed by the simple CopyLast baseline for longer prediction horizons.
	\Figure{fig:qualitative} depicts qualitative comparisons on Obj3D (\Figure{fig:qual obj}) and MOVi-A (\Figure{fig:qual movi}) between two object-agnostic baselines (ConvLSTM and PhyDNet), and our object-centric prediction framework using two distinct predictors.
	Object-agnostic baselines, which lack explicit knowledge about objects, lead to predictions of very low quality after a few time steps; whereas object-centric models, specially our OCVP-Seq, achieve accurate and temporally consistent predictions.

	\hspace{-0.5cm} \textbf{Object-Centric Evaluation:}
	We evaluate the ability of different predictors to model object dynamics.
	We convert the predicted object masks into segmentation maps by computing the \texttt{argmax} over all objects, and compare them in \Table{table:object evaluation} with ground-truth segmentation masks on MOVi-A.
	Our OCVP-Seq module achieves the best overall performance, while all transformer-based predictors outperform the LSTM module.
	\Figure{fig:segmentation} depicts frames and segmentations predicted by OCVP-Seq for an Obj3D sequence, as well as interpretable intermediate representations for three object predictions.
	We display the predicted object images and masks, and illustrate
	the corresponding relational (R.A.) and temporal attention masks.
	For dynamic and interacting objects (purple ball or green cube), relational attention focuses on neighboring and colliding objects, whereas temporal attention focuses mostly on the current input, but also attends to multiple previous time steps.
	In contrast, for static objects (purple cube), relational attention divides the attention between the corresponding object and the background, whereas temporal attention focuses only on the last time step.

	\vspace{-0.3cm}
	\begin{table}[t!]
		\centering
		\caption{OCVP-Seq Ablation Study on Obj3D. We investigate the effect of residual connections, teacher forcing (TF), skipping the slots from the first frame, and the number of predictor layers.}
		\label{table:ablation study}
		\vspace{-0.3cm}
		\setlength\tabcolsep{5.4 pt}
		\footnotesize
		\begin{tabular}{p{1.76cm} P{0.62cm}P{0.62cm}P{0.45cm}r P{0.62cm}P{0.62cm}P{0.8cm}}
			\toprule
			& \multicolumn{3}{c}{\textbf{Num Preds = 15}} && \multicolumn{3}{c}{\textbf{Num Preds = 25}} \\
			
			\textbf{} & {PSNR}$\uparrow$ & {SSIM}$\uparrow$ & {LPIPS}$\downarrow$ && {PSNR}$\uparrow$ & {SSIM}$\uparrow$ & {LPIPS}$\downarrow$ \\
			\midrule
			w/o Residual &  31.13  & 0.906 & 0.032 && 29.35 &  0.862 & 0.056 \\
			w/ Residual & \textbf{33.10} & \textbf{0.932} & \textbf{0.025} && \textbf{30.93} & \textbf{0.891} & \textbf{0.041} \\
			\midrule
			w/o TF & \textbf{33.10} & \textbf{0.932} & \textbf{0.025} && \textbf{30.93} & \textbf{0.891} & \textbf{0.041} \\
			w/ TF & 32.62 & 0.923 & \textbf{0.025} && 27.67 & 0.876 & 0.047	\\
			\midrule
			w/o Skip First & 33.10 & \textbf{0.932} & \textbf{0.025} && 30.93 & \textbf{0.891} & \textbf{0.041} \\
			w/ Skip First & \textbf{33.14} & \textbf{0.932} & \textbf{0.025} && \textbf{30.97} & \textbf{0.891} & \textbf{0.041} \\
			\midrule
			Layers = 2 & 33.04 & 0.931 & \textbf{0.025} && 30.78 & 0.889 & 0.043 \\
			Layers = 4 & \textbf{33.10} & \textbf{0.932} & \textbf{0.025} && \textbf{30.93} & \textbf{0.891} & \textbf{0.041} \\
			Layers = 6 & 32.79 & 0.929 & 0.026 && 30.73 & 0.889 & 0.043	\\
			\bottomrule
		\end{tabular}
		\vspace{-0.4cm}
	\end{table}

	\subsection{Ablation Study}
	\label{sec:ablation study}
	
	We train several modified versions of our OCVP-Seq predictor on Obj3D in order to understand how different design choices affect the video prediction performance.
	The results are reported in \Table{table:ablation study}.

	\begin{itemize}[leftmargin=0.0cm,labelsep=0.cm,label={}]
		\setlength\itemsep{0.05cm}
		
		\item  \textbf{Residual Connection:} Using a residual predictor ($\PredSlotsT{t} = \PredSlotsT{t-1} + f_{\text{pred}}( \PredSlotsT{t-1})$)
		widely outperforms its counterpart. We conclude that a residual predictor refines the input slot representations, allowing for more temporally consistent predictions.

		\item \textbf{Teacher Forcing:} When training with teacher forcing, the predictor module does not learn to handle its own imperfect predictions, thus leading to poor performance for longer prediction horizons.
		
		\item \textbf{Skip First:} Skipping the often imperfect slots from the first time step does not to help the model to predict better frames, often leading to the same SSIM and LPIPS scores.
		
		\item \textbf{Num. Layers:} Using four predictor modules leads to the best video prediction performance, especially for longer prediction horizons.
	\end{itemize}

	\section{Conclusion}
	\label{sec:conclusion}
	
	We presented a framework for object-centric video prediction,
	which decomposes video frames into object components and  models the object dynamics and their interactions to generate future video frames.
	To achieve an improved prediction performance, we proposed two OCVP transformer predictor modules, which decouple the processing of temporal dynamics and object interactions.
	In our experiments, we showed how our prediction framework using an OCVP module outperforms object-agnostic baselines, while also learning interpretable and temporally consistent object representations.
	Our models also accurately forecast the segmentation of the scene.
	In future work, we will use our rich spatio-temporal object representations for action and anticipative behavior planning.
	We release code to reproduce our results in our project website\footnote{\url{https://sites.google.com/view/ocvp-vp}}.

	\bibliographystyle{IEEEbib}
	\bibliography{referencesAngel}

\end{document}